%% file: neurips_2026.tex
\documentclass{article}

\PassOptionsToPackage{numbers, compress}{natbib}
 \usepackage[preprint]{neurips_2026}

\usepackage[utf8]{inputenc} 
\usepackage[T1]{fontenc}    
\usepackage{hyperref}       
\hypersetup{colorlinks=true,linkcolor=blue,citecolor=blue,urlcolor=blue}
\usepackage{url}            
\usepackage{booktabs}       
\usepackage{amsfonts}       
\usepackage{nicefrac}       
\usepackage{microtype}      
\usepackage{xcolor}         
\usepackage{graphicx}
\usepackage{amsmath,amssymb,booktabs,multirow,tabularx}
\usepackage{algorithm}
\usepackage{algorithmic}
\usepackage{enumitem}
 \usepackage{arydshln} 
 \usepackage{tcolorbox}
\tcbuselibrary{breakable, skins}
\usepackage{fontawesome5}   
  \usepackage{wrapfig}

\title{REVERSE: Reinforcing Evidence Verification and Search for Agentic Image geo-localization}

\author{%
  Yong Li$^{1,2,}$\thanks{Equal contribution.} \quad
  Furong Jia$^{1,}\footnotemark[1]$ \quad
  Dacheng Yin$^{3,}$\thanks{Corresponding author.} \quad
  Kang Rong$^{3}$ \\
\textbf{Fengyun Rao$^{3}$ \quad Jing LYU$^{3}$ \quad Fan Zhang$^{2,}\footnotemark[2]$} \\
  $^1$Peking University \\
  $^2$The Hong Kong University of Science and Technology \\
  $^3$WeChat Vision, Tencent Inc
}
\begin{document}

\maketitle

\begin{abstract}
Image geo-localization aims to determine where a photograph was taken, a task that often requires more than recognizing visible landmarks. Human experts typically solve it through an iterative workflow: they inspect informative regions, form location hypotheses, seek external evidence, and revise their judgments as new clues appear. Existing methods only partially capture this process: direct prediction methods bypass evidence acquisition altogether, while retrieval-augmented methods introduce external evidence but usually provide limited supervision on the intermediate decisions of where to search, how to query, and how to filter noisy results. We present REVERSE, a framework that reinforces the interplay between evidence search and verification to enable multi-turn agentic reasoning. REVERSE teaches three intermediate decisions: where to look, what to query, and what evidence to trust. To support this, we construct tool-grounded trajectories with annotated region selections, search observations, and geo-informative evidence labels, and introduce process rewards for visual grounding, query utility, and evidence discrimination. An offline search cache makes retrieval observations stable and reusable during reinforcement learning, enabling dense supervision over noisy search results. With a 4B model, REVERSE outperforms strong retrieval-augmented baselines and rivals substantially larger models on Im2GPS3k and YFCC4k. Code is available at \url{https://github.com/yonglleee/REVERSE}.
\end{abstract}

  \begin{figure}[!htbp]
      \centering
      \includegraphics[width=\linewidth]{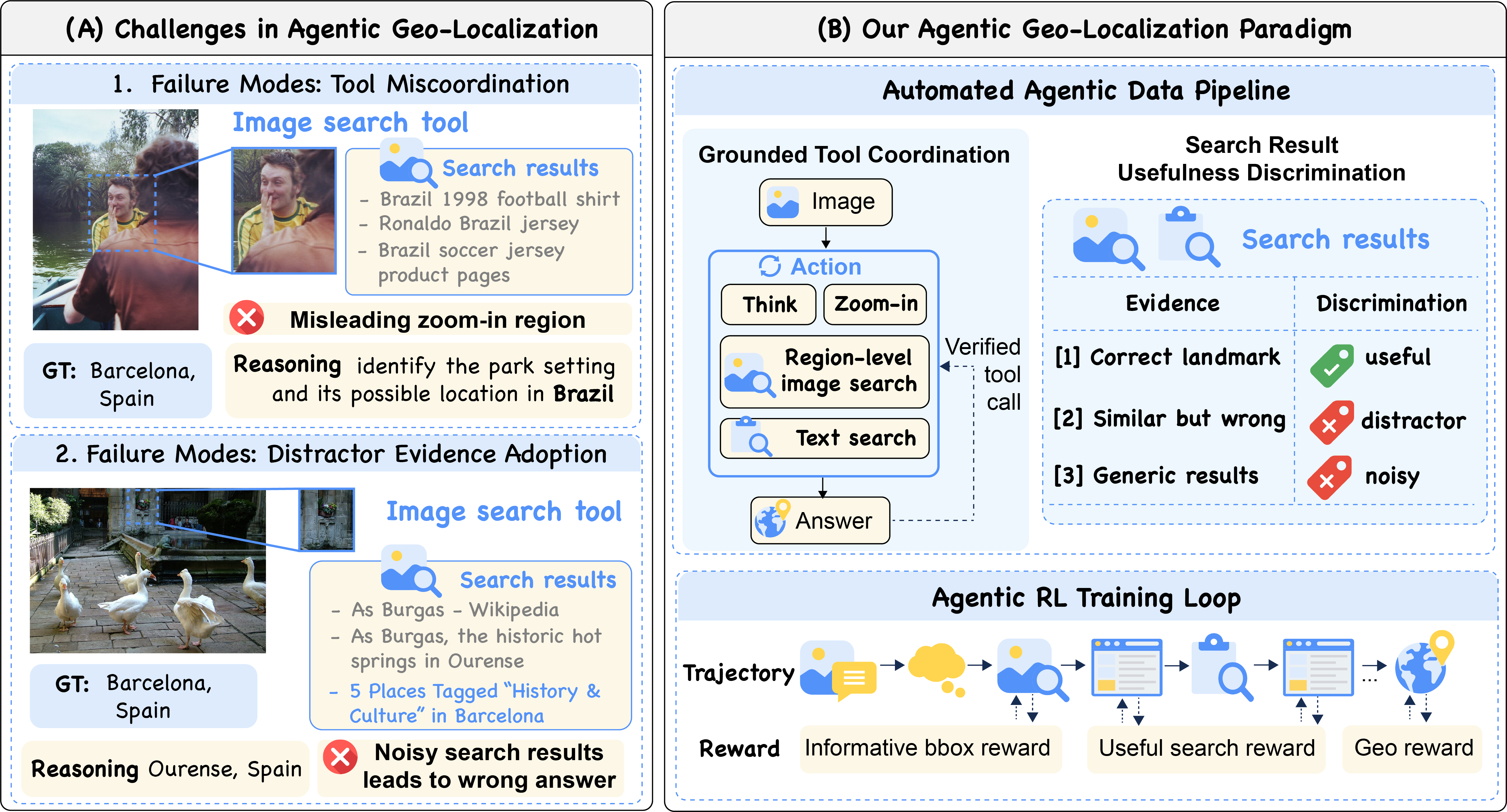}
      \caption{(A) Agentic geo-localization fails when a model crops the wrong region and follows misleading results, or retrieves the right candidate but cannot filter distractors. (B) REVERSE fixes both: spatial supervision grounds region selection, evidence labels annotate search results, and process rewards train the model via agentic RL.}
      \label{fig:title-image}
  \end{figure}

\section{Introduction}

Image geo-localization estimates the geographic location of a photograph from its
visual content~\cite{mai2024geoai,astruc2024openstreetview}, with applications
in navigation, urban analysis, and remote sensing.
Prior methods either partition the Earth into cells and train
classifiers~\cite{weyand2016planet,seo2018cplanet,muller2018geolocation},
match the image against geo-tagged databases~\cite{vivanco2023geoclip,haas2024pigeon,zhou2024img2loc},
or use large vision-language models to reason toward coordinate
predictions~\cite{jia2024g3,wu2026geor,jin2026geoagent}.
All commit to a single prediction without querying or verifying external evidence,
so performance is bounded by what the model already knows when the scene contains
unfamiliar landmarks or ambiguous visual cues.

Recent agentic methods address this by allowing models to interact with tools
across multiple turns~\cite{jia2026spotagent,ji2026thinking,han2025swarm,wang2025geovista},
but tool access alone does not guarantee reliable geo-localization.
Figure~\ref{fig:title-image}(A) shows two representative failures:
a model that crops the wrong region submits it to image search and follows
misleading results to a wrong prediction;
a model that retrieves the right candidate cannot separate it from distractors
and again predicts incorrectly.
Both failures reveal that agentic geo-localization hinges on three intermediate
decisions: \emph{where to look} (identifying which visual cues are
geographically informative), \emph{how to query} (formulating a query that
retrieves relevant external evidence), and \emph{what to trust}
(discriminating genuine evidence from visually similar distractors).
Yet none receives explicit supervision under standard final-coordinate
training, and no existing method addresses all three.

We present \textsc{REVERSE} (\textbf{R}einforcing \textbf{E}vidence
\textbf{Ver}ification and \textbf{Se}arch for Agentic Image Geo-localization),
which treats these three decisions as explicit optimization targets.
If the model must make observable choices at each step
(\emph{where to look}, \emph{how to query}, \emph{what to trust}),
each choice can be supervised directly.
We build a data pipeline that generates teacher trajectories annotated with
region selections and per-result evidence labels, providing ground truth
where final-coordinate training has none.
We design process rewards for each decision and train offline against a
pre-built search cache, so the full agentic reasoning loop is directly optimizable.

We evaluate on Im2GPS3k~\cite{vo2017revisiting} and YFCC4k~\cite{thomee2016yfcc100m}.
With a 4B model, \textsc{REVERSE} reaches 48.3\% accuracy at 25\,km on Im2GPS3k,
outperforming all retrieval-augmented baselines and matching models more than
an order of magnitude larger.

In summary, our contributions are threefold.
\begin{itemize}[leftmargin=1.5em, itemsep=0pt]
  \item 
    We introduce \textsc{REVERSE}, an agentic image geo-localization framework that reinforces evidence search and verification by jointly aligning tool use with spatial region selection, query formulation, and evidence discrimination across multi-turn reasoning chains.
  
  \item We build a tool-grounded data generation pipeline that annotates not only final locations, but also region selections, search observations, and geo-informative evidence labels, enabling supervision over where to search, how to query, and what to trust during reasoning.

  \item 
  We perform agentic reinforcement learning with a composite reward that combines geographical accuracy, format compliance, tool-use quality, and evidence discrimination, allowing a 4B vision-language model to achieve strong performance on geo-localization benchmarks.
\end{itemize}


\section{Related work}
\label{sec:related}

\subsection{Image geo-localization}
\label{sec:rw_geoloc}
Image geo-localization aims to predict the geographic location of a given image and has broad applications in urban analysis, navigation, remote sensing, and geospatial data mining.
It has long been studied as a visual recognition problem, with early methods predicting locations through geographic partitioning, hierarchical classification, or retrieval from geo-tagged image databases~\cite{weyand2016planet,seo2018cplanet,muller2018geolocation,vo2017revisiting}. Later work improves visual-location modeling with stronger backbones, and contrastive image-location alignment~\cite{pramanick2022translocator,clark2023we,vivanco2023geoclip, haas2024pigeon}. With the emergence of large vision-language models, recent methods increasingly cast geo-localization as a language-mediated reasoning problem, where visual cues can be described, compared, retrieved, and verified using external knowledge. 
Retrieval-augmented approaches ground predictions with candidate images, geographic databases, or web evidence~\cite{zhou2024img2loc,jia2024g3,jia2025georanker}, showing the value of external database beyond closed-book visual prediction. Beyond retrieval, recent LVLM-based methods further strengthen geographic reasoning through reinforcement learning~\cite{li2025recognition,wu2026geor,jin2026geoagent}.
More recent agentic methods further explore multi-agent collaboration, map-based interaction, reinforced geographic reasoning, and tool-augmented hypothesis verification~\cite{han2025swarm,ji2026thinking,jia2026spotagent}. These works demonstrate the value of external evidence and multi-step reasoning for image geo-localization, but the search and verification process itself is rarely an explicit optimization target.

\subsection{Multimodal agentic easoning.}
Large vision-language models are increasingly extended from static perception to interactive reasoning with external tools. Early tool-augmented systems connect foundation models with visual modules, program execution, web search, and structured environments, showing that complex multimodal tasks often require iterative perception, action, and evidence aggregation~\cite{yao2022react,schick2023toolformer,suris2023vipergpt,yang2023mm,wu2024v}. More recent work further internalizes such behaviors through cold start trajectories and reinforcement learning, enabling models to decide when to invoke tools, how to combine visual operations with textual search~\cite{hong2025deepeyesv2,huang2026vision,zeng2026vision,huang2025visualtoolagent,dong2026visual,zhang2025chain}. These studies indicate a broader shift from static multimodal understanding toward long-horizon evidence acquisition. However, most existing settings focus on general visual question answering or open-domain research tasks, where evidence only needs to support a textual answer. Image geo-localization requires spatially and geographically discriminative evidence for coordinate prediction, making reliable search and verification particularly challenging.

\section{Method}
\label{sec:method}

\textsc{REVERSE} recasts image geo-localization as an iterative process of visual inspection, evidence retrieval, and hypothesis verification. As illustrated in Figure~\ref{fig:title-image}, effective agentic geo-localization requires the model to make three intermediate decisions: \emph{where to look}, \emph{how to query}, and \emph{what to trust}. Our method is organized around three corresponding components: an agentic geo-localization framework that defines tool-mediated reasoning (Section~\ref{sec:agent-framework}), a data generation pipeline that constructs trajectories and evidence labels for these decisions (Section~\ref{sec:data_generation}), and a three-stage post-training procedure (SFT, Agentic Cold Start, and Agentic RL) that converts these signals into model behavior (Section~\ref{sec:training}).

\subsection{Agentic geo-localization framework}
\label{sec:agent-framework}
Expert human geo-localization is rarely a single visual matching step. A human expert first inspects potentially informative visual cues, forms tentative geographic hypotheses, searches for external evidence, and then revises the hypothesis based on the reliability of the retrieved clues. We formulate image geo-localization as a multi-turn multimodal reasoning problem that follows the same cognitive loop.
Given a query image $I$, the agent produces a trajectory
\begin{equation}
    \tau = (I, a_1, o_1, \ldots, a_T, o_T, \hat{y}),
\end{equation}
where $a_t$ denotes the action at turn $t$, $o_t$ denotes the corresponding observation, and $\hat{y}$ is the final predicted location. At each turn, the agent first generates a \texttt{<think>} trace and then either calls one tool or outputs a final answer. We restrict each response to at most one tool call so that the effect of each action can be attributed clearly during training. After a bounded number of turns, the agent must produce the final prediction in the format
\begin{equation}
    \texttt{<answer> country, city, latitude, longitude </answer>}.
\end{equation}

A central design choice is the \texttt{<useful>} tag. Search tools usually return a ranked list of results, only some of which provide reliable geographic evidence. After receiving image search or text search results, the agent is required to output a tag such as
\[
\texttt{<useful>[1, 3]</useful>},
\]
where the indices refer to retrieved results that the model judges to be geo-informative. An empty tag, \texttt{<useful>[]</useful>}, indicates that none of the returned results should be trusted. This explicit evidence selection interface makes the model's judgment observable and allows the reward to evaluate whether the agent has learned \emph{what to trust}. The agent is equipped with three tools.

\textbf{Zoom tool.}
The zoom tool takes a normalized bounding box $b=(x_1,y_1,x_2,y_2)$ and returns a cropped, resized view of the selected region. It helps the agent inspect fine-grained visual cues that may be too small or ambiguous in the full image, such as signs, road markings, inscriptions, architectural details, and distant landmarks. In our framework, zoom is not a generic enhancement operation. It is an explicit action for selecting potentially geo-informative regions.

\textbf{Image search tool.}
The image search tool performs region-level reverse image search. Instead of submitting the full image, the agent selects a bounding box and searches the cropped region. This design is important because full-image retrieval can be dominated by salient but geographically irrelevant content, while a carefully selected region can expose distinctive landmarks, storefronts, signs, or local visual patterns. The tool returns a ranked list of results with titles, source domains, and links, which the agent must then filter using the \texttt{<useful>} tag.


\textbf{Text search tool.}
The text search tool accepts a natural-language query and returns web snippets. It complements visual search by verifying candidate place names, landmarks, addresses, or textual clues inferred from previous observations. Together, the three tools cover the full reasoning loop: where to look in the image, how to query external sources, and what retrieved evidence to trust.

\subsection{Data generation pipeline}
\label{sec:data_generation}
\begin{figure}[H]
    \centering
    \includegraphics[width=\linewidth]{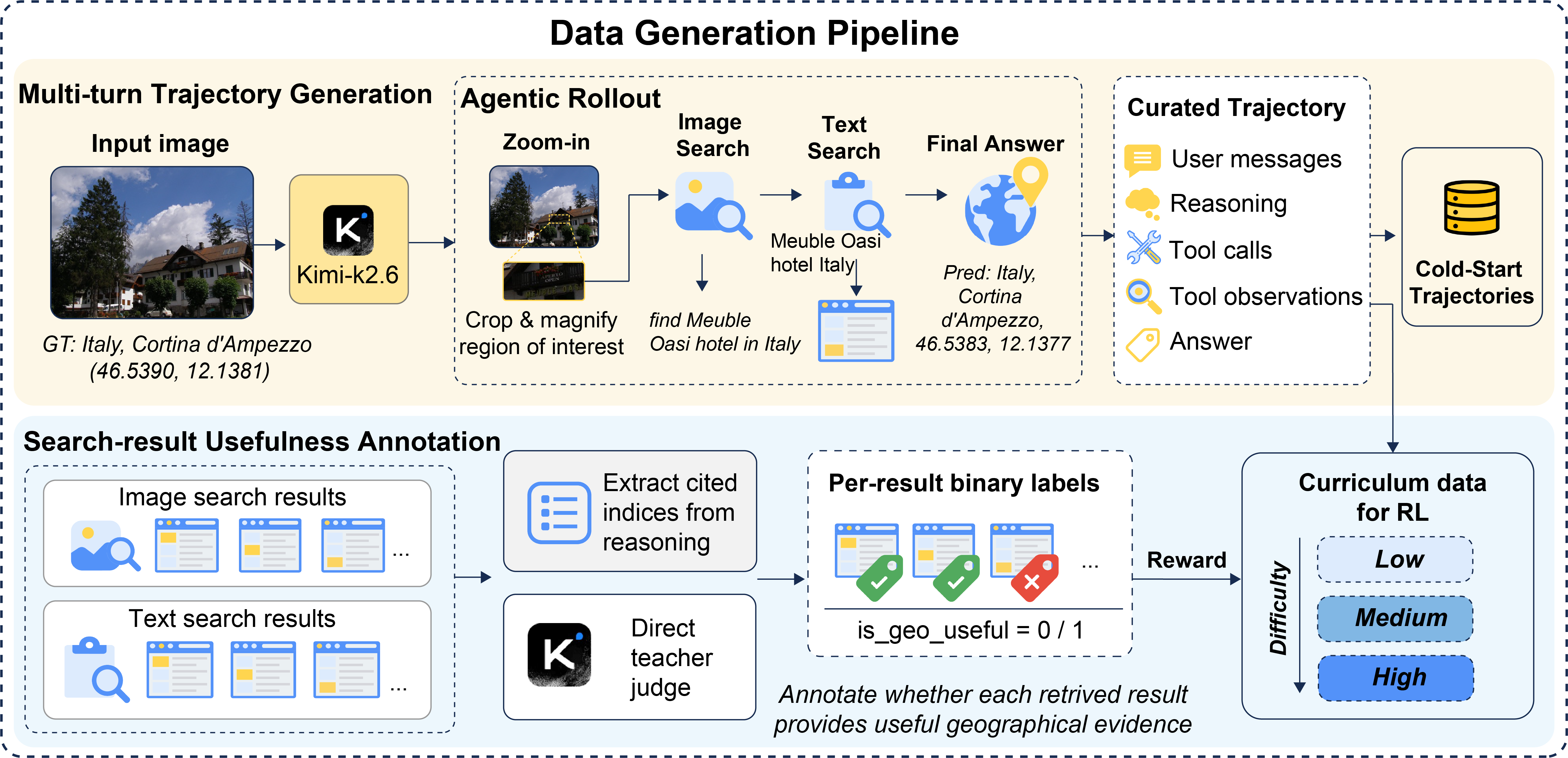}
    \caption{\textbf{Data generation pipeline.}
    Kimi-K2.6 generates multi-turn geo-localization trajectories over MP-16 Pro images using live search APIs.
    Trajectories undergo quality filtering, geo-informative label annotation, and bounding box re-annotation to correct full-image crops.
    The resulting dense annotations populate an offline cache that supports API-free RL training.}
    \label{fig:data-pipeline}
\end{figure}

Training an agent for geo-localization requires more than final coordinate labels. The environment must expose learnable signals for the full cognitive loop: \emph{where to look} through precise crop selection, \emph{how to query} through text and image search actions, and \emph{what to trust} through evidence discrimination. We construct such an environment using Kimi-K2.6 as a teacher model (Figure~\ref{fig:data-pipeline}).

\textbf{Trajectory generation and filtering.}
We source training images from the MP-16 Pro dataset and prompt Kimi-K2.6 to solve each image using the same agentic tool interface as the student model. The teacher produces multi-turn trajectories containing reasoning traces, zoom calls, region-level image search calls, text search queries, \texttt{<useful>} tags, and final answers. We filter out inaccessible images, failed teacher trajectories whose final prediction is too far from the ground truth, tool-free trajectories, and trajectories requiring excessive intervention. This yields a set of valid teacher trajectories that provide demonstrations of how visual inspection, search, and verification can be coordinated.



\textbf{Search result and crop region annotation.}
Raw teacher outputs leave two annotation gaps that we close with a second pass using Kimi-K2.6.
First, evidence labels are incomplete: for each cached image search call, the teacher judges whether each returned result provides concrete geographic evidence for the target image, using any result indices already cited in the reasoning trace as initial signals and evaluating the remaining candidates directly. This produces per-result binary labels (\texttt{is\_geo\_useful}) that later drive the evidence discrimination reward.
Second, a large fraction of image search calls use full-image bounding boxes, which weaken the \emph{where to look} signal because a model can achieve high spatial overlap without localizing any specific cue. We prompt the teacher to redraw these as tighter crops around the most geo-informative region, keeping the original search results intact. The corrected boxes serve as spatial targets for reward computation.

\textbf{Offline search cache.}
We materialize all image search and text search observations into offline caches. Each image search entry stores the annotated bounding box, geo-informative positive results, and non-informative negatives from the same ranked list---the latter are deliberately hard since they come from the same query and are often visually or semantically related to the true location. During RL rollout, model-generated image-search boxes are matched to cached entries by IoU; text queries are matched by token-level Jaccard similarity. This keeps retrieval observations stable and reusable across training steps while preserving the noise structure of real search results.

\textbf{Curriculum construction.}
The evidence discrimination reward requires every search call in a trajectory to carry usefulness labels. We retain trajectories meeting this condition, plus any that make no search calls at all, to form the full-curriculum set. A separate easy-curriculum subset restricts further to examples where the teacher's localization error is small. RL begins on this easier subset to stabilize tool-use behavior, then continues on the full-curriculum set for harder, more diverse samples.

\subsection{Three-stage training pipeline}
\label{sec:training}

\begin{figure}[H]
      \centering
      \includegraphics[width=\linewidth]{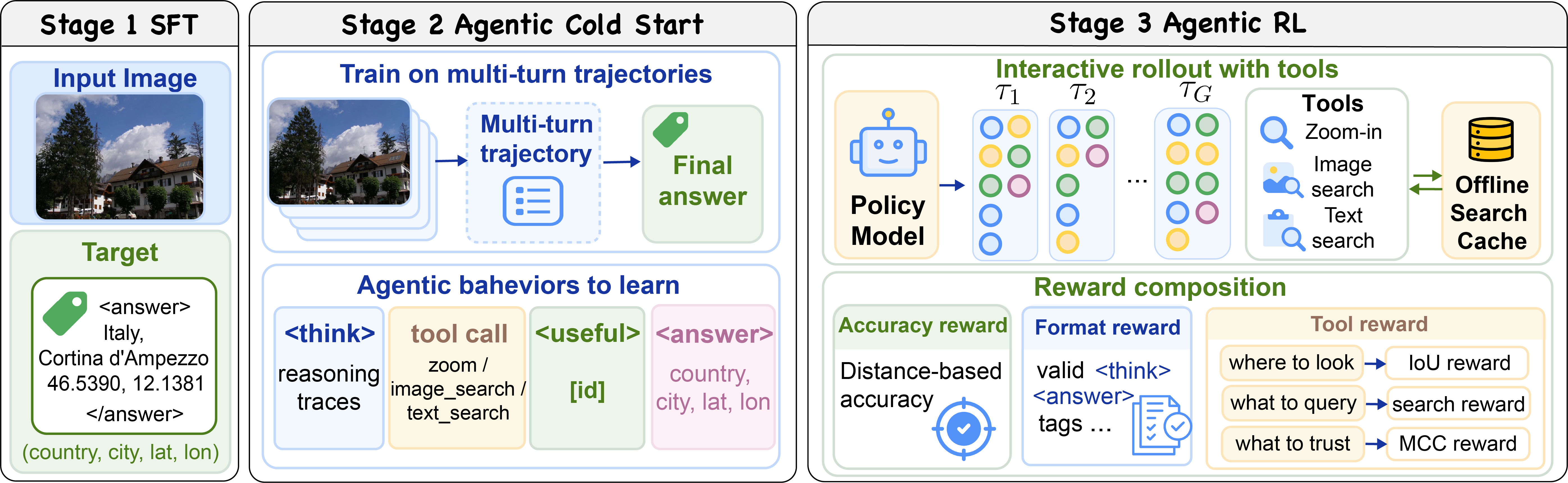}
      \caption{\textbf{REVERSE training pipeline.}
The model first learns direct image geo-localization through SFT, then acquires multi-turn tool-use behavior from cold start trajectories, and finally undergoes agentic RL with offline search observations. The reward combines geographical accuracy, format compliance, and process-level signals for where to look, how to query, and what to trust.}
      \label{fig:training-image}
  \end{figure}

As illustrated in Figure~\ref{fig:training-image}, REVERSE trains in three stages.

\textbf{Stage 1: SFT.}
We fine-tune Qwen3-VL-4B-Instruct~\cite{bai2025qwen3vl} on 4M geo-tagged images from MP16-Pro~\cite{jia2024g3},
training the model to output coordinates directly from the image with no tools or reasoning chain, the goal is to build a strong geographic memory before introducing any agentic behavior.
\begin{equation}
  \mathcal{L}_\mathrm{geo} = -\sum_{t} \log p_\theta(a_t \mid I).
\end{equation}
\textbf{Stage 2: Agentic Cold Start.}
To initialize the correct interaction protocol, we fine-tune Qwen3-VL-4B-Instruct on 4,427 curated teacher trajectories using standard autoregressive modeling:
\begin{equation}
  \mathcal{L}_\mathrm{SFT} = -\sum_{t} \log p_\theta(a_t \mid I, a_{<t}, o_{<t}).
\end{equation}
where $o_{<t}$ are the tool responses seen so far.
This stage aligns the model to the tool-call format, structured output tags, and basic evidence selection behavior before RL exploration begins.

\textbf{Stage 3: Agentic RL.}
We then optimize the cold start model with GRPO~\cite{shao2024deepseekmath}. For each image, the model samples multiple trajectories under the same tool environment and receives the composite reward defined in Section~\ref{sec:reward}. The update uses group-relative advantages, which compare rollouts generated for the same image and therefore do not require a separate critic model.  
\begin{equation}
  \mathcal{L}_\mathrm{GRPO}(\theta) = -\mathbb{E}_{\tau \sim \pi_\theta}
  \left[ A(\tau) \log \pi_\theta(\tau \mid I) \right]
  + \beta_\mathrm{KL}\, D_\mathrm{KL}(\pi_\theta \| \pi_\mathrm{ref}),
\end{equation}
where $A(\tau) = (R(\tau) - \mu_g) / \sigma_g$, $\pi_\mathrm{ref}$ is the cold start
reference policy, and $\beta_\mathrm{KL}$ keeps the policy close enough to maintain
format and tool-use stability.
RL follows the curriculum described above, moving from the easy subset to the full-coverage dataset.

\subsection{Process reward}
\label{sec:reward}

The reward is designed to align the model with both the final task objective and the intermediate decisions that make agentic geo-localization reliable. We decompose the trajectory reward into three terms:
\begin{equation}
  R(\tau) = \alpha \cdot r_\mathrm{geo} + \beta \cdot r_\mathrm{fmt} + \gamma \cdot r_\mathrm{tool},
\end{equation}
where hyperparameters $\alpha, \beta, \gamma$ control the balance between final geo-localization accuracy, strict format compliance, and intermediate tool-use quality.
Default values for all reward hyperparameters are provided in Appendix~\ref{app:hyperparams}.

\textbf{Geographical reward.}
The geographical reward $r_{\mathrm{geo}}$ evaluates the final predicted location. We compute the Haversine distance $d(\hat{y}, y)$ between the predicted and ground-truth coordinates, then map it to the standard geo-localization thresholds of 1, 25, 200, 750, and 2500 km. Predictions within these thresholds receive decreasing scores from street-level to continent-level accuracy, while parse failures and predictions beyond 2500 km receive zero reward.

\textbf{Format reward.}
The format reward $r_{\mathrm{fmt}}$ encourages stable and interpretable trajectories. A full score is assigned when the response contains valid \texttt{<think>}, \texttt{<useful>}, and \texttt{<answer>} tags in the required format. Partial credit is given when the reasoning and final answer are valid but evidence selection is missing. Invalid formatting receives zero reward.

\textbf{Tool process reward ($r_\mathrm{tool}$).}
To supervise the multi-step search process, we provide per-turn rewards that are summed and clipped to prevent reward inflation from repetitive tool usage:
\begin{equation}
  r_\mathrm{tool} = \operatorname{clip}\!\left(\textstyle\sum_{t} r_t,\; {-}0.5,\; 1.0\right).
\end{equation}
The turn-level reward $r_t$ encompasses execution rewards and a discrimination reward:
\begin{itemize}[leftmargin=1.5em, itemsep=0pt]
    \item \emph{Image search execution:} Incentivizes precise crop selection via a continuous IoU reward, $r_t^\mathrm{img} = \lambda_\mathrm{iou} \cdot \operatorname{IoU}(b_\mathrm{pred}, b_\mathrm{gt}) \cdot \mathbb{I}\!\left[\operatorname{IoU} \geq \tau\right]$.
    \item \emph{Text search execution:} Provides a fixed base reward $r_t^\mathrm{txt} = \lambda_\mathrm{base}$ per valid query.
    \item \emph{Zoom execution:} Imposes a penalty $-\delta$ for degenerate bounding boxes.
\end{itemize}

\textbf{MCC discrimination reward.}
A core challenge in training agents to consume search results is reward hacking: if the evaluation metric for evidence selection is strictly non-negative (e.g., Accuracy or F1 score), models rapidly learn to indiscriminately tag all retrieved results as \texttt{<useful>} to guarantee a positive reward.
To prevent this, we introduce a Matthews Correlation Coefficient (MCC)-based discrimination reward.
After any turn receiving search results, the model's \texttt{<useful>} tag predictions are evaluated against the geo-informative ground truth labels.
We compute MCC as:
\begin{equation}
  \mathrm{MCC} = \frac{\mathrm{TP} \cdot \mathrm{TN} - \mathrm{FP} \cdot \mathrm{FN}}
    {\sqrt{(\mathrm{TP}+\mathrm{FP})(\mathrm{TP}+\mathrm{FN})(\mathrm{TN}+\mathrm{FP})(\mathrm{TN}+\mathrm{FN})}},
\end{equation}
assigning the reward $r_t^\mathrm{mcc} = \lambda_\mathrm{mcc} \cdot \mathrm{MCC}$.
Unlike F1, MCC operates in $[-1, 1]$ and strictly penalizes both false positives and false negatives.
Crucially, it evaluates to zero for constant predictors (e.g., "mark all as useful"), completely eliminating the hacking vector and forcing the model to genuinely learn \emph{what to trust}.

\section{Experiments}
\label{sec:experiments}


We evaluate REVERSE on two standard geo-localization benchmarks: Im2GPS3k~\cite{vo2017revisiting} (2,997 geotagged images sampled worldwide) and YFCC4k~\cite{thomee2016yfcc100m} (4,536 images skewed toward everyday scenes with fewer prominent landmarks, making it harder).
We report accuracy at 1, 25, 200, 750, and 2500\,km; an unparsed prediction counts as incorrect.
Baselines span classification-based methods (PlaNet, CPlaNet, ISNs, GeoToken), retrieval-based methods (GeoCLIP, Img2Loc, PIGEON), VLM-based methods (G3, GeoBayes,Geo-R), plus untuned Qwen3-VL-4B/8B as zero-shot references.
REVERSE is built on Qwen3-VL-4B-Instruct and trained in three stages on 32 A800 GPUs; full hyperparameters are in Appendix~\ref{app:setup}.
We also conduct ablation studies on training stages, process reward components, and tool combinations.

\subsection{Main results}
\label{sec:main_results}

Table~\ref{tab:main_results} compares REVERSE to prior methods on Im2GPS3k and YFCC4k.

\textbf{Im2GPS3k.}
REVERSE achieves 48.3\% at 25\,km and 22.5\% at 1\,km on Im2GPS3k,
outperforming all prior methods at fine-grained thresholds.
Against Geo-R~\cite{wu2026geor}, the strongest comparable VLM-based method,
REVERSE gains 6.8 points at 25\,km, from 41.5\% to 48.3\%.
A 4B model trained with REVERSE also surpasses untuned Qwen3-VL-8B
at both thresholds: 22.5\% vs.\ 10.5\% at 1\,km and 48.3\% vs.\ 37.2\% at 25\,km,
showing that targeted training outweighs model scale.
At coarser thresholds, REVERSE remains competitive but trails the strongest retrieval baselines,
a gap we attribute to the near-distance bias of easy-curriculum RL
that full-curriculum training partially corrects (see \S\ref{sec:stage_ablation}).

\textbf{YFCC4k.}
On YFCC4k, REVERSE reaches 27.5\% at 25\,km, well above the untuned Qwen3-VL-4B baseline
at 12.4\%, but below retrieval-based methods such as G3 at 35.9\%.
YFCC4k skews toward landmark-scarce everyday scenes where image search returns generic or uninformative results.
Process rewards help when tools can retrieve useful evidence, but provide little leverage when the scene offers no discriminative visual cues to begin with.
This gap points to a meaningful limitation of tool-augmented approaches on visually ambiguous images.

\begin{table}[H]
  \caption{
    \textbf{Main results on Im2GPS3k and YFCC4k.}
    Accuracy (\%) at multiple distance thresholds.
    Unparsed predictions count as incorrect.
  }
  \label{tab:main_results}
  \centering
  \small
  \resizebox{\textwidth}{!}{%
  \begin{tabular}{llccccccccccc}
    \toprule
    \multicolumn{2}{c}{} & \multicolumn{5}{c}{Im2GPS3k} & \multicolumn{5}{c}{YFCC4k} \\
    \cmidrule(lr){3-7} \cmidrule(l){8-12}
    Method & Venue
    & @1km & @25km & @200km & @750km & @2500km
    & @1km & @25km & @200km & @750km & @2500km \\
    \midrule
    \multicolumn{12}{l}{\textit{\small Classification-based}} \\
    kNN                & ICCV'17    & 7.2   & 19.4  & 26.9  & 38.9  & 55.9  & 2.3   & 5.7   & 11.0  & 23.5  & 42.0  \\
    PlaNet             & ECCV'16    & 8.5   & 24.8  & 34.3  & 48.4  & 64.6  & 5.6   & 14.3  & 22.2  & 36.4  & 55.8  \\
    CPlaNet            & ECCV'18    & 10.2  & 26.5  & 34.6  & 48.6  & 64.6  & 7.9   & 14.8  & 21.9  & 36.4  & 55.5  \\
    ISNs               & ECCV'18    & 10.5  & 28.0  & 36.6  & 49.7  & 66.0  & 6.5   & 16.2  & 23.8  & 37.4  & 55.0  \\
    GeoToken           & ICDM'25    & 16.8  & 39.6  & 53.8  & 70.8  & 85.0  & \textbf{24.3}  & 35.3  & 46.6  & 64.2  & \textbf{78.6}  \\
    \multicolumn{12}{l}{\textit{\small Retrieval-based}} \\
    GeoCLIP            & NeurIPS'23 & 14.1  & 34.5  & 50.7  & 69.7  & 83.8  & 9.6   & 19.3  & 32.6  & 55.0  & 74.7  \\
    Img2Loc            & SIGIR'24   & 15.3  & 39.8  & 53.6  & 69.7  & 82.8  & 19.8  & 30.7  & 41.4  & 58.1  & 74.1  \\
    PIGEON             & CVPR'24    & 11.3  & 36.7  & 53.8  & 72.4  & 85.3  & 10.4  & 23.7  & 40.6  & 62.2  & 77.7  \\
    G3                 & NeurIPS'24 & 16.7  & 40.9  & 55.6  & 71.2  & 84.7  & 24.0  & \textbf{35.9}  & \textbf{47.0 } & \textbf{64.3 } & 78.2  \\

    \multicolumn{12}{l}{\textit{\small VLM-based}} \\
    Translocator       & ECCV'22    & 11.8  & 31.1  & 46.7  & 58.9  & 80.1  & 8.4   & 18.6  & 27.0  & 41.1  & 60.4  \\
    GeoDecoder         & ICCV'23    & 12.8  & 33.5  & 45.9  & 61.0  & 76.1  & 10.3  & 24.4  & 33.9  & 50.0  & 68.7  \\
    GeoBayes           & AAAI'26    & 6.3   & 34.7  & 53.6  & 73.7  & 85.9  & 4.9   & 16.1  & 30.9  & 55.8  & 75.4  \\
    Qwen3-VL-4B        & ---        & 8.5   & 32.3  & 47.8  & 62.6  & 71.0  & 3.0   & 12.4  & 23.2  & 38.6  & 51.9  \\
    Qwen3-VL-8B        & ---        & 10.5  & 37.2  & 55.3  & 72.8  & 84.9  & 3.3   & 10.8  & 19.6  & 31.1  & 40.4  \\
    SpotAgent          & KDD'26     & 14.1 & 40.4 & 57.8 & 73.4 & 85.8 & 7.3 & 21.5 & 36.2 & 55.0 & 70.8 \\
    Geo-R              & AAAI'26    & 18.1  & 41.5  & 58.3  & \textbf{75.3}  & \textbf{86.4}  & 10.5  & 22.7  & 40.0  & 60.8  & 75.8  \\
    \midrule
    \textbf{REVERSE (ours)} & ---   & \textbf{22.5} & \textbf{48.3} & \textbf{59.3} & 73.5 & 84.8 & 14.1 & 27.5 & 38.1 & 53.8 & 70.6 \\

    \bottomrule
  \end{tabular}%
  }
\end{table}

\subsection{Ablation studies}
\label{sec:ablation}

\textbf{Training pipeline stages.}
\label{sec:stage_ablation}
Table~\ref{tab:stage_ablation} traces each training stage on Im2GPS3k.
The base model without tools achieves 32.3\% at 25\,km; adding tools without training
slightly raises 25\,km accuracy to 36.0\% but hurts at 750\,km, from 62.6\% to 60.9\%,
confirming that tool use requires training to be effective (see also \S\ref{sec:tool_analysis}).
SFT on 4M geo-tagged images raises 25\,km accuracy by 10.2 points to 42.5\%
by building strong geographic memory before any agentic behavior is introduced.
As shown in Figure~\ref{fig:sft_scaling} in Appendix~\ref{app:sft_scaling},
accuracy improves throughout training with no sign of saturation,
and 4B and 8B models converge to nearly identical accuracy.
Agentic Cold Start transfers these priors into the agentic format:
it reaches 43.7\% at 25\,km, just above the SFT ceiling,
but coverage drops to 90.5\% as the model has not yet learned to reliably terminate.
Easy-curriculum RL restores coverage to 98\% and pushes 25\,km to 46.2\%,
stabilizing tool-use termination while continuing to improve accuracy.
Full-curriculum RL then trains on harder, long-range samples,
recovering the 750\,km and 2500\,km accuracy lost during the easy phase
and reaching 73.5\% and 84.8\% respectively.

\begin{table}[htbp]
\centering
\caption{
  \textbf{Training pipeline ablation} on Im2GPS3k.
  Each row adds one stage on top of the previous.
  \textit{Coverage}: fraction of samples producing a valid coordinate output
  (unparsed predictions count as incorrect in all accuracy columns).
  \textit{AvgTool}: average number of tool calls per sample.
}
\label{tab:stage_ablation}
\small
\resizebox{\textwidth}{!}{%
\begin{tabular}{lccccccc}
\toprule
\textbf{Stage} & \textbf{@1km} & \textbf{@25km} & \textbf{@200km} & \textbf{@750km} & \textbf{@2500km} & \textbf{Coverage} & \textbf{AvgTool} \\
\midrule
Qwen3-VL-4B (no tool)      & 8.5  & 32.3 & 47.8 & 62.6 & 71.0 & 71.3\%        & 0.0  \\
Qwen3-VL-4B (3 tools)      & 11.2 & 36.0 & 48.8 & 60.9 & 71.5 & 83.0\%        & 3.9  \\
Qwen3-VL-4B + SFT      & 18.1 & 42.5 & 57.6 & 74.8 & 85.9 & 100\%         & 0.0  \\
+ Agentic Cold Start            & 20.0 & 43.7 & 53.9 & 66.6 & 77.3 & 90.5\%        & 2.7  \\
+ Easy-curriculum RL        & 21.0 & 46.2 & 56.4 & 68.4 & 79.5 & 98.0\%        & 3.1  \\
\textbf{+ Full-curriculum RL} & \textbf{22.5} & \textbf{48.3} & \textbf{59.3} & \textbf{73.5} & \textbf{84.8} & \textbf{99.1\%} & \textbf{3.0} \\
\bottomrule
\end{tabular}%
}
\end{table}

\textbf{Process reward design.}
\label{sec:reward_ablation}
Table~\ref{tab:reward_ablation} ablates each component of the process reward.
Removing the IoU reward leaves 25\,km accuracy unchanged at 45.4\%,
suggesting the MCC signal already implicitly guides crop selection.
Removing the MCC reward drops 25\,km by 1.8 points to 43.6\%,
confirming that explicit evidence discrimination is needed to make accurate use of search results.
Removing the search utility reward causes the largest single-component drop, 2.4 points to 43.0\%,
showing it is the most critical process-level signal.
Removing all tool rewards reduces 25\,km by 3.2 points to 42.2\%,
confirming that the three components collectively contribute beyond outcome-only training.
Relaxing the IoU threshold from 0.7 to 0.5 hurts particularly at 750\,km,
from 69.8\% to 66.0\%, as the looser threshold admits off-target crops
that introduce noisy reward signal.

\begin{table}[htbp]
\centering
\caption{
  \textbf{Process reward ablation} on Im2GPS3k. All variants share the same easy-curriculum data.
  $\gamma$ = tool reward weight in the total reward; $\alpha$ = accuracy reward weight.
}
\label{tab:reward_ablation}
\small
\resizebox{\textwidth}{!}{%
\begin{tabular}{lccccccccc}
\toprule
\textbf{Variant} & \textbf{Acc} & \textbf{IoU} & \textbf{MCC} & \textbf{Search} & \textbf{$\alpha$} & \textbf{$\gamma$} & \textbf{@25km} & \textbf{@750km} & \textbf{AvgTool} \\
\midrule
Full (ours)               & \checkmark & \checkmark & \checkmark & \checkmark & 0.6 & 0.3 & \textbf{45.4} & \textbf{69.8} & 3.02 \\
w/o IoU reward            & \checkmark &            & \checkmark & \checkmark & 0.6 & 0.3 & 45.4          & 69.4          & 3.04 \\
w/o MCC reward            & \checkmark & \checkmark &            & \checkmark & 0.6 & 0.3 & 43.6          & 69.0          & 3.0 \\
w/o Search reward         & \checkmark & \checkmark & \checkmark &            & 0.6 & 0.3 & 43.0          & 68.0           & 3.02  \\
Acc only  & \checkmark &            &            &            & 1.0 & 0   & 42.2          & 67.8          & 2.90 \\
\midrule
IoU thresh 0.5  & \checkmark & \checkmark & \checkmark & \checkmark & 0.6 & 0.3 & 43.4          & 66.0          & 3.15 \\
\bottomrule
\end{tabular}%
}
\end{table}




\subsection{Analysis}

\textbf{Tool selection.}
\label{sec:tool_analysis}
Table~\ref{tab:tool_ablation} evaluates which tools contribute to accuracy
using the same trained REVERSE model on Im2GPS3k.
Image search is the dominant tool: used alone, it reaches the highest accuracy at most thresholds
by retrieving visually similar landmarks to anchor location hypotheses.
Text search adds complementary gains at fine-grained thresholds,
grounding the model's reasoning with web evidence.
Combining both achieves the best balance across thresholds,
and adding zoom on top yields the best overall results,
consistent with the IoU reward having trained effective crop selection.
Full results are in Table~\ref{tab:tool_ablation} (Appendix~\ref{app:tool_ablation}).



\begin{figure}[H]
      \centering
      \includegraphics[width=\linewidth]{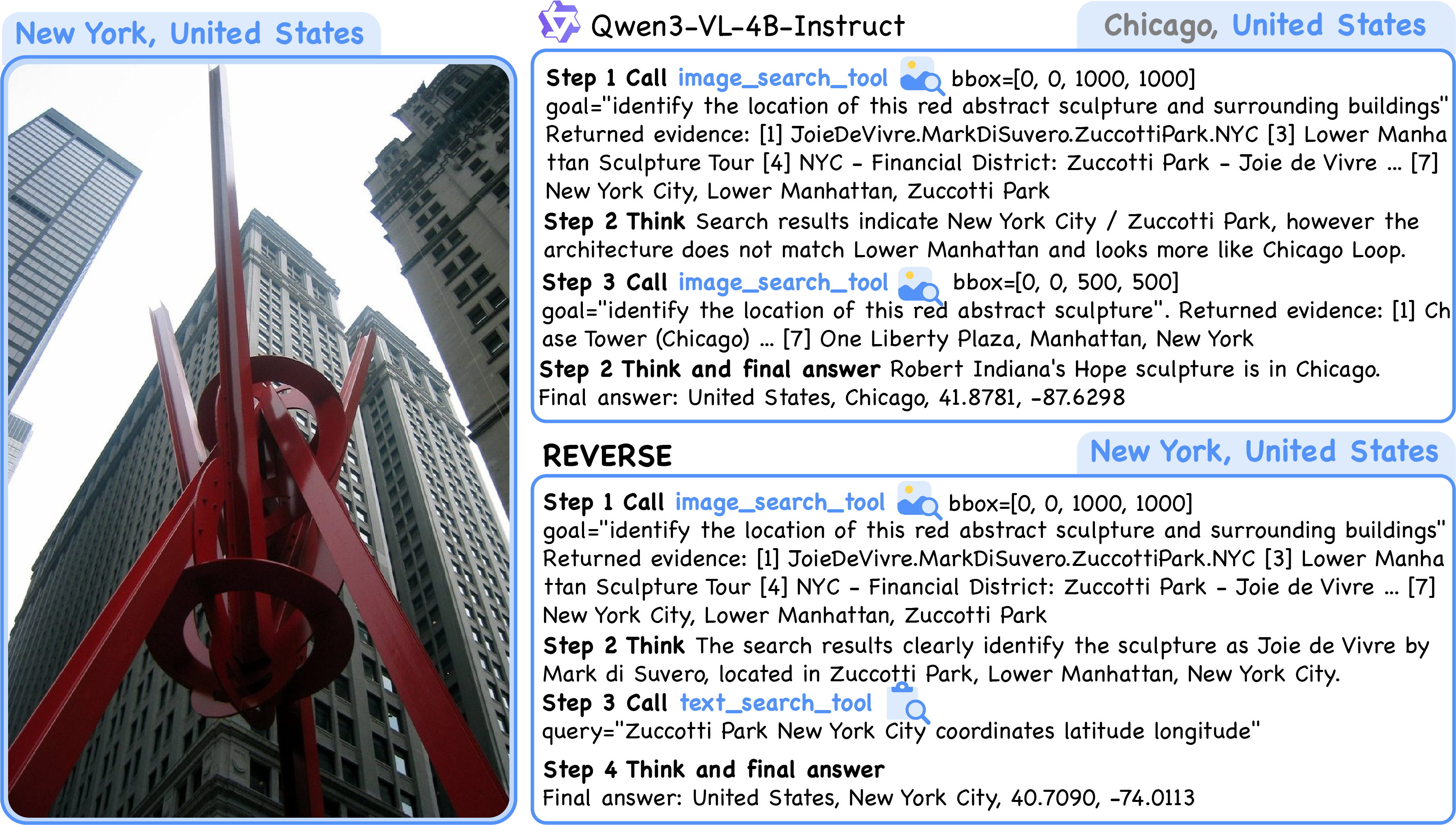}
      \caption{\textbf{Comparison of two agent trajectories on the same geo-localization case.} The Qwen3-VL-4B-Instruct trajectory first retrieves strong evidence for Joie de Vivre at Zuccotti Park, New York City, but rejects it based on a Chicago architectural prior and follows a cropped-search distractor to Chicago. \textbf{REVERSE}'s trajectory keeps the full-image evidence, identifies the sculpture as Mark di Suvero’s Joie de Vivre in Zuccotti Park, and correctly predicts New York City.}
      \label{fig:case-image}
  \end{figure}

\textbf{Case study.}
\label{sec:qualitative}
Figure~\ref{fig:case-image} shows a case where REVERSE succeeds while untuned Qwen3-VL-4B-Instruct fails.
Both models receive identical image search results pointing clearly to Zuccotti Park, New York City.
The untuned model rejects this evidence, reasoning that the architecture looks more like Chicago Loop,
and follows a second search to the wrong city.
REVERSE correctly reads the first result, identifies the sculpture as Joie de Vivre in Zuccotti Park,
and uses text search to confirm coordinates before answering.
The contrast illustrates the core failure mode of untuned VLMs as tool-users:
strong visual priors override retrieved evidence even when the evidence is unambiguous.

\begin{wrapfigure}{r}{0.5\columnwidth}
  \centering
  \vspace{-6pt}
  \includegraphics[width=0.45\columnwidth]{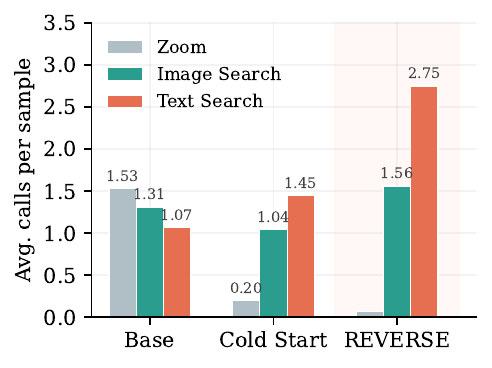}
  \caption{\textbf{Tool calls per sample} across training stages on Im2GPS3k.
  Base: untuned Qwen3-VL-4B. Cold Start: Stage~2. REVERSE: full-curriculum RL.}
  \label{fig:tool_dist}
  \vspace{-8pt}
\end{wrapfigure}

\textbf{RL shapes tool-use patterns, not just frequency.}

Figure~\ref{fig:tool_dist} compares per-tool invocation rates across three training stages.
The untuned model over-relies on zoom at 1.53 calls per sample and invokes all three tools
indiscriminately, with no apparent sense of when each is useful.
Agentic Cold Start cuts zoom usage to 0.20 and shifts toward text search at 1.45,
reflecting that the SFT trajectories teach the image-search to text-search reasoning chain.
REVERSE sharpens this further: zoom nearly disappears at 0.07,
image search rises to 1.56, and text search reaches 2.75 calls per sample.
This progressive shift from visual-only to search-grounded reasoning is exactly
the behavior our process rewards are designed to elicit, and the numbers confirm it.




\section{Conclusion}
\label{sec:conclusion}

This paper addresses image geo-localization by recasting it as an active, multi-turn reasoning process.
The key idea is that a model should learn to inspect regions, retrieve external evidence,
and judge what to trust, rather than committing to a single prediction from one forward pass.
We train a 4B VLM with zoom and search tools via GRPO, supervised by process-level rewards
that cover where to search, how to query, and what to believe.
On Im2GPS3k, REVERSE reaches 48.3\% at 25\,km, outperforming prior VLM-based methods
and surpassing the untuned 8B baseline despite using a model half its size.
A core finding is that tool use must be trained, not just enabled:
equipping untuned models with tools hurts accuracy, while process rewards teach
the model to use tools selectively and interpret their results critically.


\bibliographystyle{unsrtnat}
\bibliography{references}
\newpage

\appendix

\input{appendix.tex}



\end{document}

%% file: appendix.tex
%
%

\tcbset{
  promptbox/.style={
    enhanced,
    arc=3pt,
    boxrule=0.7pt,
    colframe=black,
    colbacktitle=black!85,
    coltitle=white,
    fonttitle=\small\sffamily\bfseries,
    titlerule=0pt,
    toptitle=3pt,
    bottomtitle=3pt,
    left=8pt, right=8pt, top=6pt, bottom=6pt,
    fontupper=\small,
    colback=gray!6,
  },
}

\section{Agent Implementation Details}
\label{app:impl}

\subsection{Prompt}
\label{app:system_prompt}

To ensure consistency between training and evaluation, we adopt a unified prompt
template used verbatim during both Agentic Cold Start and RL rollout.
The template is designed to guide the model through systematic visual reasoning
of geographic clues while enforcing a \emph{think-then-act} pattern for tool use:
the model must always reason in \texttt{<think>} before issuing any tool call or
final answer.
Beyond standard tool-use instructions, our prompt introduces two REVERSE-specific
mechanisms: (i) explicit \textbf{decision rules} that govern which tool to call
given the visual context (e.g., image search for landmarks, zoom for small text),
and (ii) an \textbf{evidence discrimination} tag \texttt{<useful>} that forces the
model to identify which retrieved results are geographically relevant to the
specific image---this tag directly drives our MCC process reward during RL training.

\subsection{Example Inference Trace}
\label{app:example_trace}

The trace below shows a complete two-turn inference from REVERSE on a real
Im2GPS3k image (id:~311938754, prediction error: 3.4\,km).
The model identifies a distinctive astronomical instrument via image search,
discriminates the relevant results with \texttt{<useful>}, and confirms the
exact coordinates through a follow-up text search before issuing its final answer.

\subsection{Agent Tool Infrastructure}
\label{app:tools}

REVERSE exposes three tools to the agent at inference time.
All tool calls are formatted as structured JSON inside
\texttt{<tool\_call>~\ldots~</tool\_call>} tags, following the same format
used during Agentic Cold Start training.

\textbf{Image Zoom Tool.}
The zoom tool implements a deterministic crop-and-resize pipeline.
Given normalized bounding-box coordinates $[x_1, y_1, x_2, y_2]$ in
$[0, 1000]^2$, the tool maps them to pixel-level crops via linear scaling
to the image dimensions.
The crop is then resized following Qwen3-VL's \texttt{smart\_resize} scheme:
output resolution is snapped to multiples of the patch factor (28 pixels),
with a minimum of $256{\times}256$ and a maximum of $2048{\times}1024$ pixels,
preserving aspect ratio.
The resized crop is re-injected into the model's context as a new image token
sequence, allowing fine-grained inspection of text, inscriptions, or small
visual details not visible at full scale.

\textbf{Image Search Tool.}
The reverse image search tool crops the specified region using the same pipeline
as the zoom tool, uploads the crop to cloud object storage to obtain a public URL,
and queries the Oxylabs Google Lens API.
Results are returned as a ranked list of up to 10 web pages, each containing
a title, source URL, and domain.
The agent marks useful results with \texttt{<useful>[\ldots]</useful>} tags,
which serve as the discrimination signal for the MCC reward during RL training.

\textbf{Web Search Tool.}
The text search tool submits natural-language queries to the Tavily Search API,
returning up to 5 results per query with titles, snippets, and source URLs.
The tool supports parallel multi-query execution: if the agent provides a list
of queries, all are dispatched concurrently and results are concatenated.
API keys are managed via a pool with automatic rotation on quota exhaustion.
Both image search and text search results are cached in a local SQLite database
(WAL mode) to ensure reproducibility across evaluation runs.

\subsection{Offline Search Cache for RL Rollout}
\label{app:cache}

Live API calls during RL rollout are expensive and non-reproducible.
We pre-annotate the training corpus with tool call--response pairs offline and
cache them in a Parquet-based lookup table keyed by (image hash, bounding box)
for image search and by query string for text search.
During rollout, cache hits are served instantly; cache misses fall back to live
API calls.
This hybrid strategy reduces API cost by over 90\% while retaining full coverage
for novel queries generated by the evolving policy.

\begin{tcolorbox}[promptbox, title={Prompt Template}]
You are a geolocation expert. Given an image, identify its location.

\medskip
\textbf{Available Tools}

The agent can invoke the following tools:
\begin{itemize}[leftmargin=1.5em, itemsep=3pt, topsep=3pt]
  \item \textbf{image\_search\_tool}: Reverse image search using a cropped region.
    Best for distinctive landmarks, buildings, or scenes. Returns matching web pages.
    \begin{itemize}[leftmargin=1.2em, itemsep=1pt, topsep=1pt]
      \item[\textit{--}] \textit{Parameters}: \texttt{bbox\_2d [x1, y1, x2, y2]} --- crop region;
        \texttt{goal} --- description of what to identify.
    \end{itemize}
  \item \textbf{text\_search\_tool}: Search the web with natural language queries.
    Use for visible text/signs, landmark names, or any clues found from image search results.
    Supports a list of parallel queries.
    \begin{itemize}[leftmargin=1.2em, itemsep=1pt, topsep=1pt]
      \item[\textit{--}] \textit{Parameters}: \texttt{query} --- a string or list of search strings.
    \end{itemize}
  \item \textbf{image\_zoom\_in\_tool}: Zoom into a region to read text or inscriptions
    that are too small at full scale.
    \begin{itemize}[leftmargin=1.2em, itemsep=1pt, topsep=1pt]
      \item[\textit{--}] \textit{Parameters}: \texttt{bbox\_2d [x1, y1, x2, y2]} --- region to zoom.
    \end{itemize}
\end{itemize}

\medskip
\textbf{Decision Rules}
\begin{itemize}[leftmargin=1.5em, itemsep=2pt, topsep=2pt]
  \item Distinctive landmark or scene visible (but uncertain of exact coords)
    $\to$ use \textbf{image\_search\_tool}.
  \item Text/signs already legible $\to$ use \textbf{text\_search\_tool} directly.
  \item Text/signs too small to read $\to$ use \textbf{image\_zoom\_in\_tool} first,
    then \textbf{text\_search\_tool}.
  \item Image search returns a landmark/location name $\to$ follow up with \textbf{text\_search\_tool}.
  \item Do \textbf{not} use \textbf{image\_zoom\_in\_tool} before \textbf{image\_search\_tool}
    --- zoom does not improve image search.
  \item When in doubt, use a tool. Only skip tools when certain.
  \item \textbf{Fallback rule}: If tools have failed, returned empty results, or exceeded
    call limits, do \textbf{not} keep retrying. Provide your best \texttt{<answer>}
    based on visual priors. Never finish a response without a \texttt{<tool\_call>}
    or \texttt{<answer>}.
\end{itemize}

\medskip
\textbf{Reasoning Requirement}

For every response, first enclose your reasoning in \texttt{<think>~\ldots~</think>} tags,
then output \textbf{exactly one} of:

\smallskip
{\small\ttfamily
\texttt{<tool\_call>\{"name": "image\_search\_tool", "arguments": \{"bbox\_2d": [x1,y1,x2,y2], "goal": "..."\}\}</tool\_call>}\\[2pt]
\texttt{<tool\_call>\{"name": "text\_search\_tool", "arguments": \{"query": "..."\}\}</tool\_call>}\\[2pt]
\texttt{<tool\_call>\{"name": "text\_search\_tool", "arguments": \{"query": ["q1","q2"]\}\}</tool\_call>}\\[2pt]
\texttt{<tool\_call>\{"name": "image\_zoom\_in\_tool", "arguments": \{"bbox\_2d": [x1,y1,x2,y2]\}\}</tool\_call>}
}

\smallskip
or your final \texttt{<answer>} when confident without needing a tool.

\medskip
\textbf{Evidence Discrimination}

After receiving tool results, output on its own line:
\texttt{<useful>[i, j, \ldots]</useful>}
listing the 1-based indices of results that match this specific image.
Results about a different place are \textbf{not} useful even if they contain
geographic information. Output \texttt{<useful>[]</useful>} if none match.

\medskip
\textbf{Final Answer Format}

When ready to answer, provide the final answer inside \texttt{<answer>~\ldots~</answer>} tags.
The final answer must include Country, City, Latitude, and Longitude (decimal format).\\
Example: \texttt{<answer>Italy, Golfo Arnaci, 40.9606, 9.5873</answer>}

Latitude and longitude must always be valid numeric values and cannot be null.
\end{tcolorbox}


\begin{tcolorbox}[promptbox, title={Inference Trace --- Im2GPS3k id: 311938754 \quad GT: Jaipur, India \quad Error: 3.4\,km}]

\textbf{User:} \textit{[Image: circular hemispherical astronomical instrument, stone construction, Jaipur, India]}

\medskip\hrule\smallskip

\textbf{Assistant (Turn 1):}

\texttt{<think>}
The image shows a circular structure with a reflective surface, appearing to be
a sundial or astronomical instrument. This is a distinctive landmark.
Let me run image search to identify it.
\texttt{</think>}

\texttt{<tool\_call>\{"name": "image\_search\_tool", "arguments": \{"bbox\_2d": [0, 0, 1000, 1000], "goal": "identify this circular astronomical instrument"\}\}</tool\_call>}

\smallskip\hrule\smallskip

\textbf{Tool Response (\texttt{image\_search\_tool}):}

\textbf{[1]} File:JaiPrakashYantraJaipur20080213-1.jpg --- Wikimedia Commons\\
\textbf{[2]} Jai Prakash Yantra, measuring time through a hemispherical\ldots\ --- traveladventures.org\\
\textbf{[3]} Jantar Mantar, Jaipur --- Wikipedia\\
\textbf{[4]} Detail of the Jai Prakash Yantra, a sundial which measures\ldots\ --- Alamy\\
\textbf{[5]} Jantar Mantar Observatory --- markandchucksadventures.com\\
\textbf{[6]} File:Zodiac Circle jantar mantar, Jaipur 1.jpg --- Wikimedia Commons\\
\textbf{[7]} Jantar Mantar: Jaipur's Famous Observatory --- throughmylens.com\\
\textbf{[8]} Jai Prakash Yantra Jantarmantar Observatory --- Tripadvisor\\
\textbf{[9]} Under Jaipur Skies: Jantar Mantar --- shaopeng.blog\\
\textbf{[10]} 617 Sundial Jantar Mantar Jaipur Stock Photos --- Dreamstime.com

\smallskip\hrule\smallskip

\textbf{Assistant (Turn 2):}

\texttt{<think>}
The results clearly identify this as the Jai Prakash Yantra at Jantar Mantar
Observatory in Jaipur, Rajasthan, India.
Let me confirm the exact coordinates with a text search.
\texttt{</think>}

\texttt{<useful>[1, 2, 3, 4, 5]</useful>}

\texttt{<tool\_call>\{"name": "text\_search\_tool", "arguments": \{"query": "Jantar Mantar Observatory Jaipur coordinates latitude longitude"\}\}</tool\_call>}

\smallskip\hrule\smallskip

\textbf{Tool Response (\texttt{text\_search\_tool}):}

\textbf{[1]} GPS coordinates of Jantar Mantar (Jaipur), India.
Latitude: 26.9215, Longitude: 75.8213 --- latitude.to\\
The Jantar Mantar monument of Jaipur is a collection of nineteen astronomical instruments
built by the Rajput king Sawai Jai Singh, completed in 1738\,CE.
Latitude: 26°55'17.39''N, Longitude: 75°49'16.79''E. Nearest city: Jaipur.\\
\textbf{[2]} Jantar Mantar, Jaipur --- Wikipedia\\
Kranti Vritta Yantra (measures longitude and latitude of celestial bodies)\ldots
Its face is angled at 27 degrees, the latitude of Jaipur.\\
\textbf{[3]} Jantar Mantar Jaipur India --- tourhq.com\\
The Jantar Mantar features the world's largest stone sundial and is a UNESCO World Heritage site.\\
\textbf{[4]} Jantar Mantar --- banbanjara.com\\
Sawai Jai Singh II constructed the astronomical observatory in Jaipur.
19 astrological and architectural instruments including the Vrihat Samrat Yantra.\\
\textbf{[5]} UNESCO Portal to the Heritage of Astronomy\\
The Jantar Mantar of Jaipur is an observatory built in the first half of the 18th century.
One of the most complete collections of pre-telescopic masonry instruments in the world.

\smallskip\hrule\smallskip

\textbf{Assistant (Final Answer):}

\texttt{<think>}
Confirmed: Jantar Mantar Observatory, Jaipur, India.
Coordinates 26.9215°N, 75.8213°E from the first result.
\texttt{</think>}

\texttt{<useful>[1]</useful>}

\texttt{<answer>India, Jaipur, 26.9215, 75.8213</answer>}

\end{tcolorbox}

\section{Training Data Construction}
\label{app:data}

Our training corpora are built from oracle-generated trajectories using
Kimi-K2.6 as the annotation agent.
We randomly sample 100K images from MP16-Pro and prompt Kimi-K2.6 to solve
each using the same agentic tool interface as REVERSE, without revealing the
ground-truth location.
This yields $\sim$30,750 valid raw trajectories after discarding failed or
inaccessible cases.
The remaining MP16-Pro images (approximately 4M) are used for Stage~1
SFT as coordinate-only supervision.
We apply a cascade of quality filters to the 30,750 trajectories to produce
training splits of varying strictness (Table~\ref{tab:data_stats}).

\paragraph{Trajectory filtering.}

\emph{RL base} (19,962 trajectories): retain samples where the image is
accessible, the oracle's localization error $d \leq 200$\,km, at least
one tool was invoked, the trajectory did not exceed the maximum turn budget,
and no tool call returned an API failure response.
This is the broadest usable set and forms the basis for all downstream splits.

\emph{Agentic Cold Start} (4,427 trajectories, subset of RL base): additionally
require at most 5 tool calls per trajectory.
This stricter constraint keeps only the most concise, self-consistent
reasoning paths for supervised fine-tuning.

\emph{RL full-coverage} (6,996 trajectories, subset of RL base): retain
samples where every image-search call in the trajectory has been annotated
with geo-informative and non-informative result labels, and at least one
positive result exists per search call.
This annotation is required so that the MCC discrimination reward
$r_t^\mathrm{mcc}$ can be computed for every tool call during RL training.
Trajectories with no search calls are also included.

\emph{Easy-curriculum subset} (3,104 trajectories, subset of RL full-coverage):
further restrict to $d \leq 25$\,km to form the first-stage RL curriculum,
focusing training on geographically unambiguous scenes before progressing to
harder long-range samples.

\paragraph{Offline cache construction.}
The image search cache contains 273K entries indexed by (image path, call index).
Each entry stores the oracle's annotated bounding box, all geo-informative
(positive) results, and all non-informative (negative) results from the same
search call.
During RL rollout, a model-generated bounding box is matched to a cache entry
if $\operatorname{IoU}(b_\mathrm{pred}, b_\mathrm{gt}) \geq \tau$; otherwise
the tool returns an empty result, penalizing imprecise crops.
The text search cache contains 82K entries indexed by a normalized query hash,
matched at rollout time by token-level Jaccard similarity.
Both caches cover all training and validation images, enabling fully offline RL
training with no live API calls.

\begin{table}[h]
\centering
\small
\caption{Dataset statistics for Agentic Cold Start and RL training.}
\label{tab:data_stats}
\begin{tabular}{lcl}
\toprule
\textbf{Split} & \textbf{Samples} & \textbf{Filter criteria} \\
\midrule
Agentic Cold Start        & 4,427  & $d \leq 200$\,km, $\leq$5 tool calls \\
RL easy-curriculum    & 3,104  & full-coverage + $d \leq 25$\,km \\
RL full-curriculum    & 6,996  & full-coverage, $\geq$1 positive result per search call \\
\bottomrule
\end{tabular}
\end{table}

\section{Training Hyperparameters}
\label{app:hyperparams}

Table~\ref{tab:hyperparams} summarizes the hyperparameters for all three training stages.
REVERSE follows a progressive training schedule: Stage~1 establishes broad geographic
priors via large-scale supervised learning; Stage~2 fine-tunes the model on agentic
tool-use trajectories at a lower learning rate; Stage~3 applies GRPO-based RL with a
two-phase curriculum, starting from easy samples ($d \leq 25$\,km) before expanding
to the full difficulty range.

\begin{table}[h]
\centering
\caption{
  \textbf{Training hyperparameters for all three stages of REVERSE.}
}
\label{tab:hyperparams}
\resizebox{\textwidth}{!}{%
\begin{tabular}{lccc}
\toprule
\textbf{Hyperparameter} & \textbf{Stage 1 (SFT)} & \textbf{Stage 2 (Agentic Cold Start)} & \textbf{Stage 3 (Agentic RL)} \\
\midrule
Base model              & Qwen3-VL-4B-Instruct & Stage 1 checkpoint & Stage 2 checkpoint \\
Training framework      & FSDP                 & FSDP               & VERL + GRPO        \\
Rollout engine          & ---                  & ---                & SGLang             \\
Nodes ($\times$ 8 GPUs) & 8                    & 1                  & 4                  \\
\midrule
Training data           & MP16-Pro (4M)        & Cold-start V5.2 (4,427) & Easy (3,104) $\to$ Full (6,996) \\
Batch size              & 512                  & 128                & 256                \\
Learning rate           & $2\times10^{-5}$     & $2\times10^{-5}$   & $1\times10^{-6}$   \\
LR schedule             & cosine               & cosine             & cosine             \\
Warmup ratio            & 0.1                   & 0.1                & 0.05               \\
Min-LR ratio            & ---                  & ---                & 0.1                \\
Training steps          & $\approx$7,800       & 35                 & 80 (easy) + 135 (full) \\
\midrule
Reward weights          & ---                  & ---                & $\alpha{=}0.6,\,\beta{=}0.1,\,\gamma{=}0.3$ \\
IoU reward coeff        & ---                  & ---                & 0.2                \\
MCC reward coeff        & ---                  & ---                & 0.3                \\
Base search reward      & ---                  & ---                & 0.1                \\
IoU threshold           & ---                  & ---                & 0.7                \\
Max turns (rollout)     & ---                  & ---                & 10                 \\
Max response length     & ---                  & ---                & 16,384 tokens      \\
\midrule
Hardware                & \multicolumn{3}{c}{8 $\times$ NVIDIA A800 80\,GB per node} \\
\bottomrule
\end{tabular}%
}
\end{table}

\begin{table}[h]
\centering
\small
\caption{Reward hyperparameters and their default values.}
\label{tab:reward_params}
\begin{tabular}{llc}
\toprule
\textbf{Symbol} & \textbf{Meaning} & \textbf{Default} \\
\midrule
$\alpha$ & Geolocation reward weight & 0.6 \\
$\beta$  & Format reward weight      & 0.1 \\
$\gamma$ & Tool reward weight        & 0.3 \\
\midrule
$\tau$              & IoU gate threshold           & 0.7 \\
$\lambda_\mathrm{iou}$  & Image search IoU coeff   & 0.2 \\
$\lambda_\mathrm{base}$ & Text search base reward  & 0.1 \\
$\lambda_\mathrm{mcc}$  & MCC discrimination coeff & 0.3 \\
$\delta$            & Zoom invalid-box penalty     & 0.05 \\
\bottomrule
\end{tabular}
\end{table}

\section{Experimental Setup}
\label{app:setup}

\textbf{Benchmarks.}
We evaluate on two standard geolocation benchmarks.
\emph{Im2GPS3k}~\cite{vo2017revisiting} contains 2,997 geotagged images sampled worldwide,
covering diverse geographic regions and landmark types.
\emph{YFCC4k}~\cite{thomee2016yfcc100m} contains 4,536 images from the YFCC100M dataset
and skews toward everyday scenes with fewer prominent landmarks, making it harder than Im2GPS3k.
Following standard practice, accuracy at threshold $r$ counts a prediction as correct
if its Haversine distance to ground truth is within $r$\,km; unparsed predictions count as incorrect.
We report accuracy at 1, 25, 200, 750, and 2500\,km.

\textbf{Baselines.}
\label{app:baselines}
We compare against a broad range of prior geolocation methods.
\emph{Classification-based} methods partition the Earth into geographic cells and train classifiers:
PlaNet~\cite{weyand2016planet}, CPlaNet~\cite{seo2018cplanet}, ISNs~\cite{muller2018geolocation},
kNN~\cite{vo2017revisiting}, and GeoToken~\cite{ghasemi2025geotoken}.
\emph{Retrieval-based} methods match query images against geo-tagged databases:
GeoCLIP~\cite{vivanco2023geoclip}, Img2Loc~\cite{zhou2024img2loc}, and PIGEON~\cite{haas2024pigeon}.
\emph{VLM-based} methods leverage pretrained vision-language models with geographic supervision:
Translocator~\cite{pramanick2022translocator}, GeoDecoder~\cite{clark2023geodecoder},
G3~\cite{jia2024g3}, and GeoBayes~\cite{shi2026geobayes}.
We additionally evaluate Qwen3-VL-4B and Qwen3-VL-8B~\cite{bai2025qwen3vl} without any
geographic fine-tuning as untuned VLM baselines.
\emph{RL-based} methods optimize geographic reasoning with reinforcement learning:
Geo-R~\cite{wu2026geor}.

\textbf{Implementation details.}
REVERSE is built on Qwen3-VL-4B-Instruct~\cite{bai2025qwen3vl} and trained in three stages (full hyperparameters in Table~\ref{tab:hyperparams}).
\textbf{Stage 1 (SFT):} We fine-tune on 4M geo-tagged images from MP16-Pro~\cite{jia2024g3} for $\approx$7,800 steps on 8 nodes with batch size 512 and learning rate $2\times10^{-5}$.
\textbf{Stage 2 (Agentic Cold Start):} We fine-tune on 4,427 curated teacher trajectories for 35 steps on 1 node with batch size 128 and learning rate $2\times10^{-5}$.
\textbf{Stage 3 (Agentic RL):} We apply GRPO~\cite{shao2024deepseekmath} with a two-phase curriculum: 80 steps on 3,104 easy samples ($d \leq 25$\,km) followed by 135 steps on 6,996 full-coverage samples, using learning rate $1\times10^{-6}$ and reward weights $\alpha{=}0.6$, $\beta{=}0.1$, $\gamma{=}0.3$.
RL rollout uses VERL~\cite{sheng2024hybridflow} and SGLang~\cite{zheng2023sglang}.
All training runs on NVIDIA A800 80\,GB GPUs (8 GPUs per node).

\section{Tool Combination Results}
\label{app:tool_ablation}

\begin{table}[h]
\centering
\caption{
  \textbf{Tool combination ablation} on Im2GPS3k.
  All variants use the same REVERSE model (full-curriculum RL).
  Unparsed predictions count as incorrect.
}
\label{tab:tool_ablation}
\small
\begin{tabular}{lccccc}
\toprule
\textbf{Tools} & \textbf{@1km} & \textbf{@25km} & \textbf{@200km} & \textbf{@750km} & \textbf{@2500km} \\
\midrule
No tool                     & 8.5  & 32.3 & 47.8 & 62.6 & 71.0 \\
Zoom only                   & 12.0 & 38.0 & 58.0 & 64.0 & 84.0 \\
Image search only           & 20.0 & 48.0 & 64.0 & 72.0 & 76.0 \\
Text search only            & 14.0 & 42.0 & 54.0 & 64.0 & 76.0 \\
Zoom + text search          & 12.0 & 38.0 & 58.0 & 68.0 & 76.0 \\
Image + text search         & 22.0 & 47.0 & 58.0 & 72.0 & 80.0 \\
\textbf{Image + text + zoom} & \textbf{22.5} & \textbf{48.3} & \textbf{59.3} & \textbf{73.5} & \textbf{84.8} \\
\bottomrule
\end{tabular}
\end{table}

\section{SFT Scaling}
\label{app:sft_scaling}

\begin{figure}[h]
  \centering
  \includegraphics[width=\linewidth]{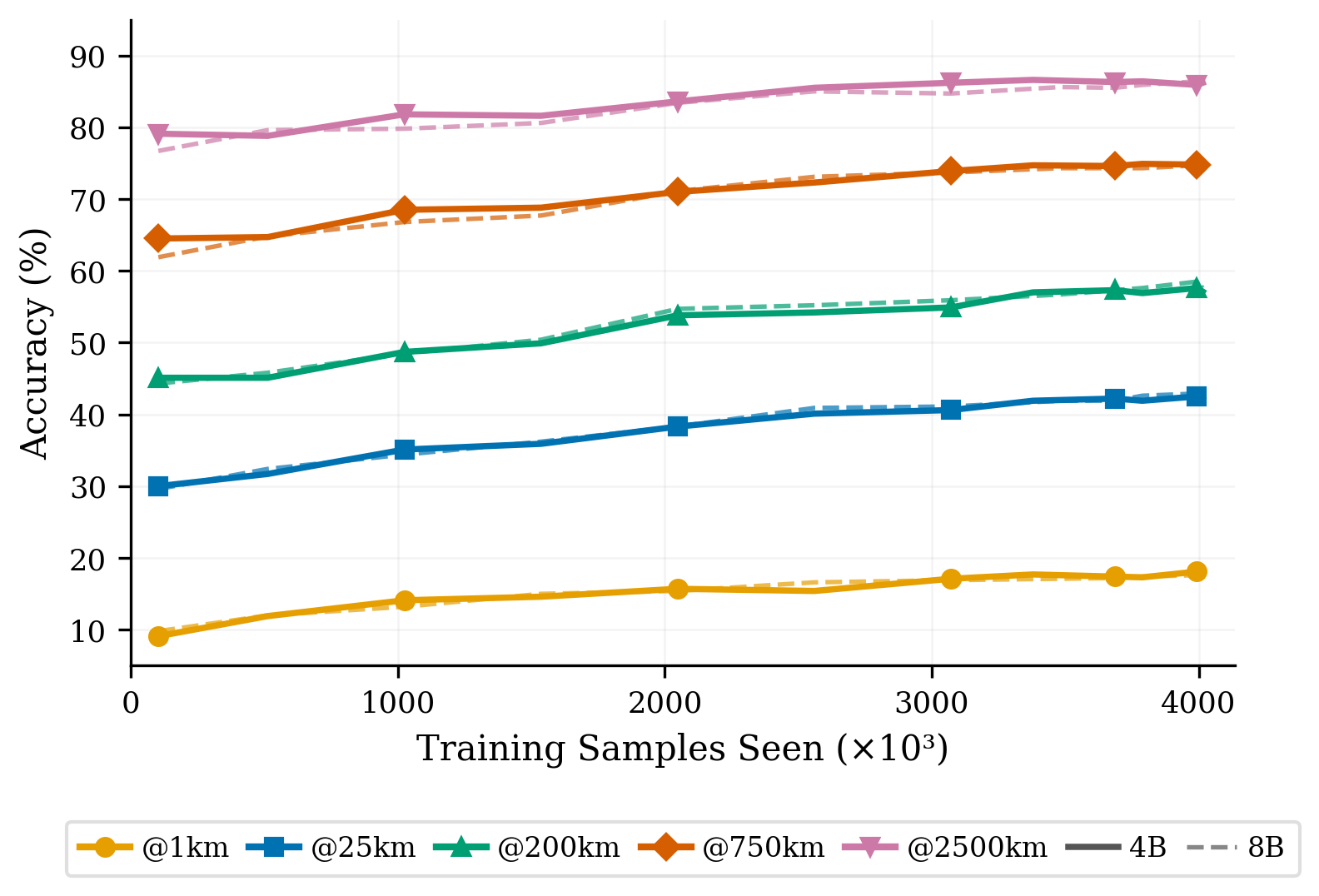}
  \caption{
    \textbf{SFT scaling curves} on Im2GPS3k (2,997 images).
    Accuracy at five distance thresholds as a function of training samples seen.
    Solid lines: Qwen3-VL-4B; dashed lines: Qwen3-VL-8B.
  }
  \label{fig:sft_scaling}
\end{figure}

Figure~\ref{fig:sft_scaling} shows the accuracy of Qwen3-VL-4B and Qwen3-VL-8B
on Im2GPS3k as a function of training samples seen during SFT (Stage~1).
In this stage, the model is trained to directly output coordinates without any
intermediate reasoning or tool use, purely consolidating geographic visual memory.
Both models improve consistently across all distance thresholds with no sign of
saturation at 4M samples.
Notably, 4B and 8B converge to nearly identical accuracy, with 8B providing only
marginal gains ($<$1\,pp at @25km), motivating our choice of the 4B model.

\section{Limitations and Broader Impacts}
\label{app:limitations}

REVERSE trades live retrieval for reproducibility: RL training uses a static offline cache,
which enables high-throughput training and controlled evaluation but means the model
learns search behaviors against a fixed snapshot rather than a shifting live web.
Similarly, agentic cold-start trajectories come from Kimi-K2.6; while this provides high-quality
supervision, it introduces a dependency on a large proprietary teacher.
We also note that our image search tool does not filter domains that expose GPS metadata
such as Flickr, so retrieved results could in principle contain ground-truth coordinates.
We did not observe leakage artifacts, but explicit domain filtering is a straightforward
improvement for future work.

Accurate geolocation agents raise privacy concerns: they could identify individuals or
sensitive locations from casually shared photographs.
REVERSE is a research system exploring spatial reasoning and multi-turn tool use,
and real-world deployment should include safeguards such as metadata stripping
and blurring of faces and private spaces.